\documentclass{article}

     \PassOptionsToPackage{numbers, compress}{natbib}

     \usepackage[final]{neurips_2020}

\usepackage[utf8]{inputenc} 
\usepackage[T1]{fontenc}    
\usepackage{hyperref}       
\usepackage{url}            
\usepackage{booktabs}       
\usepackage{amsfonts}       
\usepackage{nicefrac}       
\usepackage{microtype}      
\usepackage[pdftex]{graphicx}
\usepackage{latexsym}
\usepackage{soul}
\usepackage{float}
\usepackage{amsmath}
\usepackage{multirow}
\usepackage{subcaption}
\usepackage{color}
\usepackage{xstring}
\usepackage{tabularx}
\usepackage{natbib}
\renewcommand{\vec}[1]{\mathbf{#1}}

\usepackage[prependcaption,textsize=scriptsize]{todonotes}
\definecolor{mycolor}{HTML}{FF6600}
\definecolor{benjicolor}{HTML}{0000FF}
\definecolor{kayodecolor}{HTML}{FF0000}

\newcolumntype{C}{>{\centering\arraybackslash}X}
\newcolumntype{L}{>{\raggedright\arraybackslash}X}
\newcolumntype{R}{>{\raggedleft\arraybackslash}X}

\title{Towards Localisation of Keywords in Speech \\ Using Weak Supervision}

\author{%
  Kayode Olaleye\thanks{Correspondence: kaykola.olaleye@gmail.com} \qquad Benjamin van Niekerk \qquad Herman Kamper \\
  Department of E\&E Engineering \\
  Stellenbosch University  
}

\begin{document}

\maketitle

\begin{abstract}
Developments in weakly supervised and self-supervised models could enable speech technology in low-resource settings where full transcriptions are not available. We consider whether keyword localisation is possible using two forms of weak supervision where location information is not provided explicitly. In the first, only the presence or absence of a word is indicated, i.e.\ a bag-of-words (BoW) labelling. In the second, visual context is provided in the form of an image paired with an unlabelled utterance; a model then needs to be trained in a self-supervised fashion using the paired data. For keyword localisation, we adapt a saliency-based method typically used in the vision domain. We compare this to an existing technique that performs localisation as a part of the network architecture. While the saliency-based method is more flexible (it can be applied without architectural restrictions), we identify a critical limitation when using it for keyword localisation. Of the two forms of supervision, the visually trained model performs worse than the BoW-trained model. We show qualitatively that the visually trained model sometimes locate semantically related words, but this is not consistent. While our results show that there is some signal allowing for localisation, it also calls for other localisation methods better matched to these forms of weak supervision.
\end{abstract}

\section{Introduction}

There is a growing body of work considering how speech processing systems can be developed in the absence of conventional transcriptions~\citep{synnaeve2014,duong2016,palaz2016, settle2017,weiss2017}. Several of these studies consider the setting where we have an indication of whether a word occurs in an utterance or not, but don't know where the word occurs. Given this weak form of supervision, we ask whether it is still possible to localise words in a speech utterance.

We specifically consider two types of weak supervision. In the first, speech audio is paired with a bag-of-words (BoW) labelling, indicating the presence or absence of a word without giving the location, order, or number of occurrences~\citep{palaz2016}. This is useful when 
only noisy labels are available, e.g., for low-resource languages. In the second, images are paired with spoken captions. This form of visual supervision can be useful when it is not possible to collect textual labels, e.g., for languages without a written form. 
Since both the utterance and the paired image are unlabelled, some form of self-supervision is required, where a proxy task is used to train the model~\citep{doersch+zisserman_iccv17,aytar+vondrick+torralba_neurips16, ngiam+ng_icml11,owens+wu+torralba_eccv16,wang+gupta_iccv}.
Existing approaches include using an external image tagger
to extract soft multi-class labels~\citep{pasad2019,kamper2019a}, or a model can be trained so that the images and the speech are projected into a joint embedding space~\citep{harwath2016,chrupala2017}. Compared to full transcriptions, both  forms of supervision are closer to the signals that infants would have access to while learning their first language~\citep{pinker1994,roy2003,chrupala2016,eimas+quinn94,bomba1983,boves2007}, and to how one 
would teach new words   
to robots using spoken 
language~\citep{meng2013}.
Moreover, these weak forms of supervision
could conceivably be easier to obtain when developing systems for low-resource languages~\citep{de1998}.

We consider two localisation mechanisms. The first was introduced in~\citep{palaz2016}, referred to as PSC based on the author names. PSC uses a convolutional neural network (CNN) architecture to jointly locate and classify words using BoW supervision. Here we extend PSC by also considering visual supervision. The second is GradCAM~\citep{selvaraju2017}, a saliency-based method originally built for locating objects in images using the gradients of a target concept with respect to filter activations. Here we apply GradCAM for localisation of keywords in speech with both forms of supervision.

We find that PSC-based localisation outperforms GradCAM. However, the underlying model used by GradCAM performs better on word \textit{detection} (the task of identifying whether a word is present in an utterance or not without considering location). We speculate that this is due to a mismatch between the multi-label classification loss used here and the activation estimation method of GradCAM, which was originally developed for single-label multi-class classification. Unsurprisingly, we find that BoW-trained models outperform visually trained models on the localisation task. We show qualitatively that this is sometimes caused by the visual supervision capturing semantic and other information from the scene not matching the spoken caption exactly. However, although this aligns with previous studies on other tasks~\citep{chrupala2017,kamper2019b}, this finding is not always consistent. Taken together, our results suggest that other self-supervision and localisation methods need to be considered that are better matched to the form of multi-label weak supervision used here.

\begin{figure}[t]
    \centering
	\includegraphics[width=0.95\linewidth]{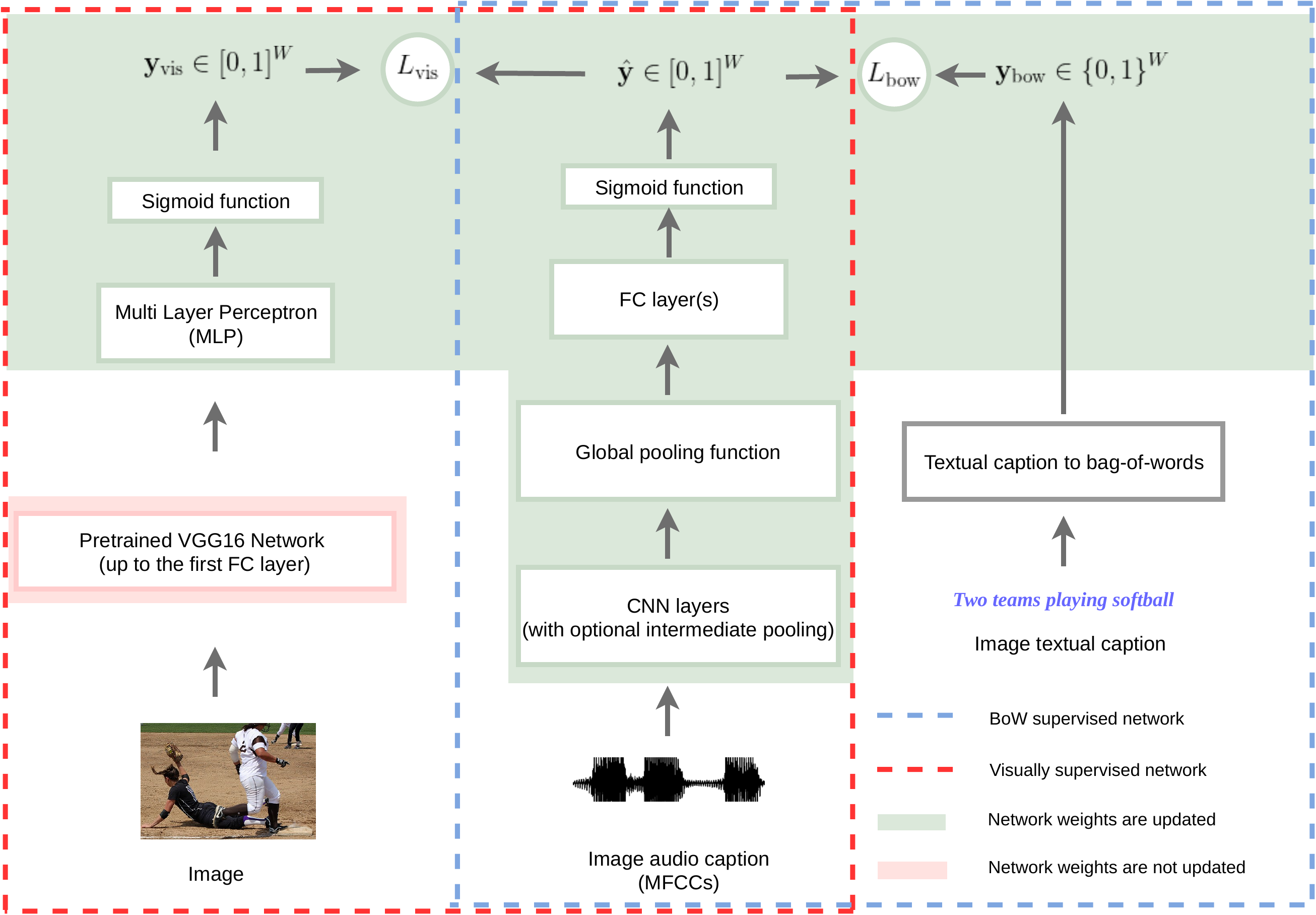}
	\caption{We consider localisation with networks trained with bag-of-word (blue, right) and visual (red, left) supervision.}
	\label{fig:arch}
\end{figure}

\section{Proposed Localisation Mechanisms and Models}

\paragraph{Two forms of weak supervision.}
In the absence of full transcriptions, \citet{palaz2016} considered a weak form of supervision where each speech sequence $X = (\vec{x}_1, \cdots, \vec{x}_T)$ is paired with BoW labels $\vec{y}_{\textrm{bow}} \in \left \{0, 1 \right \}^W$ ({blue} dotted region on the right in Figure~\ref{fig:arch}). Each element $y_{\textrm{bow},w}$ is an indicator for whether word $w$ occurs in the utterance. The utterance $X$ consists of $T$ acoustic vectors (Mel-frequency cepstral coefficients 
in our case), and $W$ is the total number of words in the vocabulary.

In another form of weak supervision, each utterance $X$ is paired with a corresponding image $I$ (red region, Figure~\ref{fig:arch} left). In our case $I$ is a scene and $X$ is a spoken caption \citep{harwath2015,harwath2016,harwath2017}. \citet{kamper2019b} proposed that, in an attempt to get the same type of supervision as in the BoW case,  $I$ can be passed through a trained multi-label image tagger. The tagger is trained on external data, and can therefore be seen as a way to utilise existing vision systems to obtain a noisy target which can be used for self-supervision. 
Concretely, the tagger outputs soft probabilities  $\vec{y}_{\textrm{vis}} \in \left [0, 1 \right ]^W$ for whether a particular word is relevant to the image. We use the same visual tagger as in \citep{kamper2019b}.

\begin{figure}[t]
	\centering
	\includegraphics[width=1.0\columnwidth]{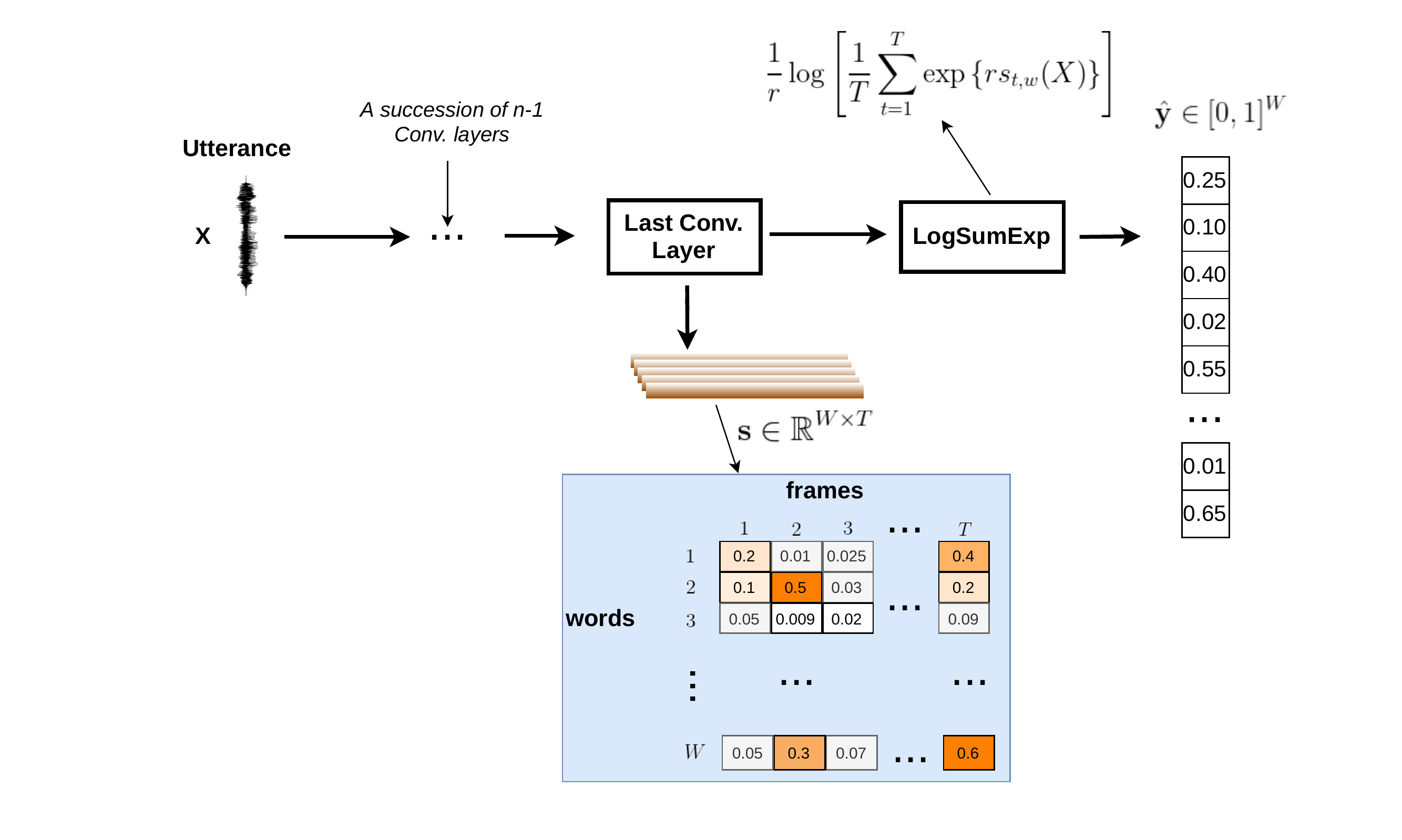}
	\captionof{figure}{An illustration of the PSC model. The blue region represents how the output of the last convolutional layer is unpacked to give a score $s_{t,w}$ for each frame $t$ and each word $w$.
    In this example, the
    proposed location for the 
    keywords corresponding to indexes $2$ and $W$ would respectively be 
    frames $2$ and $T$, which achieve the highest scores in the depicted 
    blue region.}
	\label{fig:psc_architecture}
\end{figure}

\paragraph{The PSC model.}

The PSC model is designed to simultaneously perform detection and localisation of keywords in  speech utterances when provided with BoW targets~\citep{palaz2016}. The structure of the model is illustrated in Figure~\ref{fig:psc_architecture}. The model consists of a number of convolutional layers which operates on $X$ to produce a score for the presence or absence of a word at a particular time-step. Concretely, the model's last one-dimensional convolutional layer produces filter outputs $\mathbf{s}_1, \mathbf{s}_2, \ldots \mathbf{s}_T$, where each $\mathbf{s}_t \in \mathbb{R}^W$ is a $W$-dimensional vector. We therefore have one convolutional filter per word, with ${s}_{t, w}$ giving the score for word $w$ at time $t$ (depicted as 
the blue region in 
Figure~\ref{fig:psc_architecture}). These frame-level localisation scores are fed into an aggregation function to produce a single utterance-level detection score:
\begin{equation}
	g_w(X) = \frac{1}{r} \log \left[ \frac{1}{T} \sum^T_{t = 1} \exp \left\{r s_{t, w}(X)\right\} \right]
\end{equation}
This LogSumExp aggregation function is equivalent to average pooling when $r \rightarrow 0$ and max pooling when $r \rightarrow \infty$. According to \citet{palaz2016}, this intermediate aggregation operation drives the weights of frames which have similar scores close to each other during training, resulting 
in better localisation performance. The final output of the network is the probability that each word is present in the utterance: $\hat{\vec{y}} = \sigma ( \vec{s}(X))$, with $\sigma$ the sigmoid function. The model is trained with the summed binary log loss:
\begin{equation}
	L(\hat{\mathbf{{y}}}, \mathbf{y}_\text{bow}) = -\sum_{w=1}^W \left\{ y_{\text{bow},w} \log \hat{y}_w + (1 - y_{\text{bow},w}) \log [1 - \textrm{log} \hat{y}_w] \right\}
	\label{eq:multilabel}
\end{equation}
where $\vec{y}_{\text{bow}}$ indicates that BoW supervision is used. For visual supervision, it is replaced by $\vec{y}_{\text{vis}}$. Note that in this model, the localisation mechanism is built into the model by directly connecting the last convolutional layer with the set of words. If we had any other layers between the scoring layer and the output, this connection would be lost.

\paragraph{The GradCAM model.}
In contrast to PSC, GradCAM is a method that can be used for localisation in any CNN architecture~\citep{selvaraju2017}. However, here we apply it within a particular model. We therefore refer to both the localisation mechanism and our specific CNN architecture together as the ``GradCAM model'' (although these are technically disjoint). Our GradCAM model again has a number of convolutional layers, but also uses intermediate max-pooling layers, max-pooling over time over its last convolutional layer, and a number of fully connected layers before terminating in its output (see Section~\ref{sect:models}).
The output is again denoted as $\hat{\mathbf{y}} \in [0, 1]^W$ and the same loss~\eqref{eq:multilabel} as for PSC is used. After training the model, GradCAM localisation is performed as follows. Let us call the  output from the last one-dimensional convolutional layer $\mathbf{q}$, i.e., it produces $\mathbf{q}_1, \mathbf{q}_2, \ldots, \mathbf{q}_T$, where each $\mathbf{q}_t \in \mathbb{R}^K$ is a $K$-dimensional vector  with $K$ the number of filters.
(In the PSC model, the number of filters in the last convolutional layer had to be $W$, but we do not have
this constraint here.) We first determine the ``importance" of the $k^{\textrm{th}}$ filter to the word $w$:
\begin{equation}
\gamma_{k, w} = \frac{1}{T} \sum^T_{t = 1} \frac{\partial \hat{y}_w}{\partial q_{t, k}}
\end{equation}
This value indicates how closely the $k^{\textrm{th}}$ convolutional filter pays attention to word $w$, based on the trained model weights. For a particular utterance, we want to know the scores $s_{t, w}$ of word $w$ at time $t$. We obtain this by calculating the values of all the filters, and then weighing each filter by its importance $\gamma_{k, w}$. We are only interested in changes that would result in a higher output score for a word, so we take the $\textrm{ReLU}$, resulting in $s_{t, w} = \textrm{ReLU} \left[ \sum_{k = 1}^K \gamma_{k, w} q_{t, k} \right]$.

\section{Experimental Setup and Evaluation}

\label{ssec:data}
We use the Flickr8k Audio Caption Corpus of~\citet{harwath2015}, consisting  of five English audio captions for each of the roughly $8$k images. $30$k utterances are used for training, $5$k for development, and $5$k for testing. To obtain BoW labels for the training data, we construct indicator vectors $\vec{y}_\text{bow} \in \left \{0, 1 \right \}^W$ for the $W = 1000$ most common words in the transcriptions. For visual supervision, each training image is passed through the multi-label visual tagger of \citet{kamper2019b}, which uses VGG-16~\citep{simonyan2014} and is trained on images~\citep{imagenet2009,krizhevsky2012,girshick2014} disjoint from the data considered here. The result is soft labels $\vec{y}_{\textrm{vis}} \in \left [0, 1 \right ]^W$ for the $W = 1000$ image classes from the tagger.

\label{sect:models}
Our PSC model consists of six one-dimensional convolutional layers with ReLU activations. The first has $96$ filters with a kernel width of $9$ frames. The next four has a width of $11$ units. The last convolutional layer, with $W = 1000$ filters and a width of $11$ units, is fed into the LogSumExp aggregation function with a final sigmoid activation. Our GradCAM model consists of three one-dimensional convolutional layers with ReLU activations. Intermediate max pooling over 3 units are applied in  the first two layers. The first convolution has $64$ filters with a width of $9$ frames. The second layer has $256$ filters with a width of $11$ units, and the last layer has $1024$ filters with a width of $11$. Global max pooling is applied
followed by a sigmoid activation to obtain the final output for the $W = 1000$ words. All models are 
implemented in PyTorch and uses Adam optimisation \citep{kingma2014} with a learning rate of $1\cdot{10}^{-4}$.

\begin{figure}[t]
	\centering
	\includegraphics[width=1.0\columnwidth]{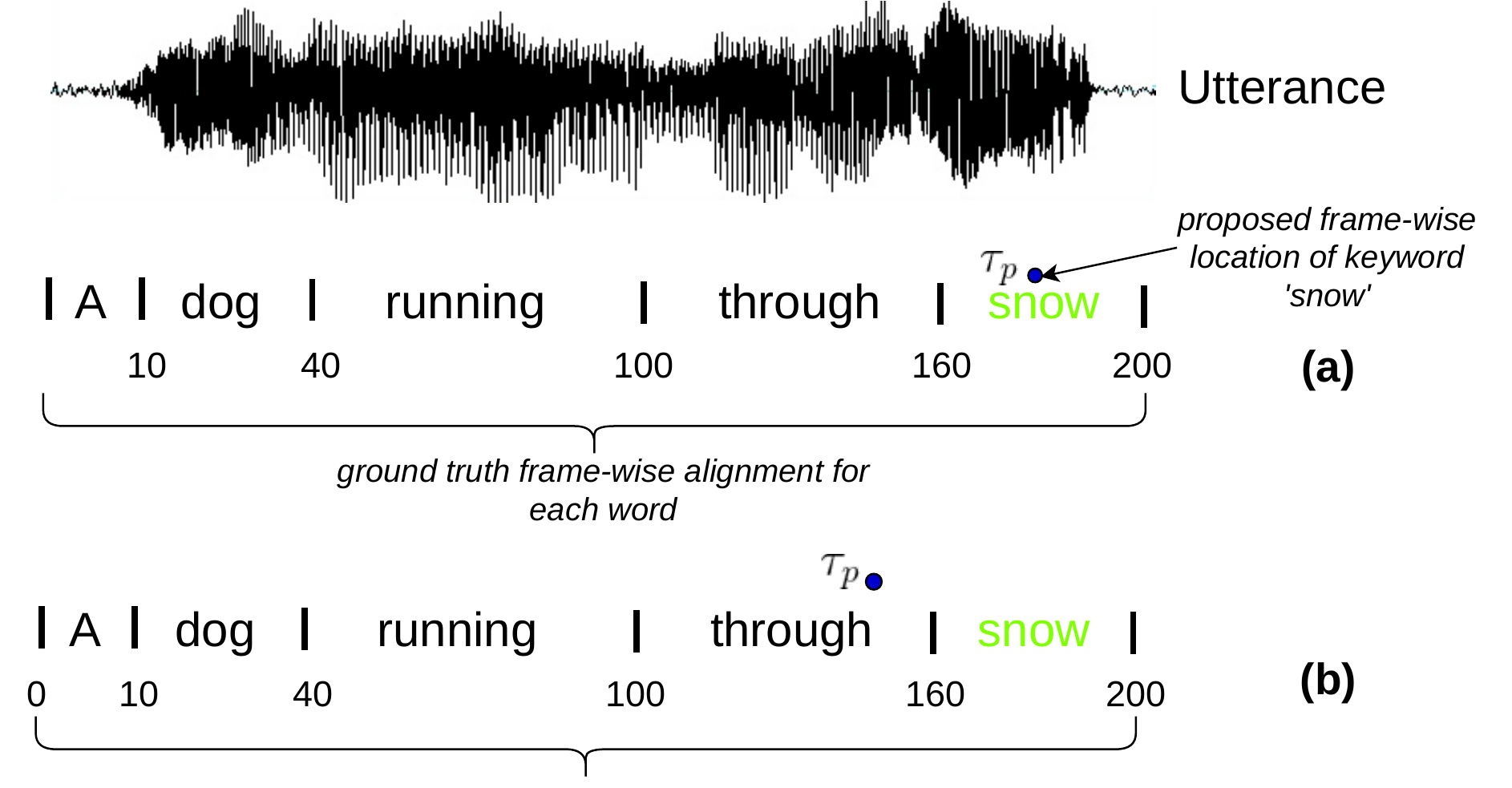}
	\vspace{5pt}
	\captionof{figure}{Illustration of the evaluation procedure: (\textbf{a}) depicts a successful localisation attempt of the keyword in green (\textit{snow}), while (\textbf{b}) depicts a failed attempt.}
	\label{fig:evaluation_procedure}
\end{figure}

\label{ssec:eval}
As in~\citep{palaz2016}, we consider two settings for evaluating localisation performance. In the \textit{oracle} setting, we assume that the system perfectly detects whether a word occurs in an utterance or not. We then evaluate whether it is also able to locate it.
The position of the highest score is taken as the predicted location: 
$\tau_p = \text{argmax}_t(s_{t,w})$.
$\tau_p$ is accepted as correct
if the frame $\tau_p$ is within the interval corresponding to the true $w$, according to forced alignments. Figure~\ref{fig:evaluation_procedure} depicts two localisation attempts 
by a model. 
The proposed location for the keyword \textit{snow} is $\tau_p$ and its ground truth location spans frames $160$ and $200$. In (\textbf{a}), $\tau_p$ is accepted as a correct location of \textit{snow} because it is within the ground truth frames, and rejected in (\textbf{b}) because it is outside the ground truth frames. In the \textit{actual} evaluation setting, detection is also taken into account. Before evaluating localisation, we apply a threshold $\alpha$ to the detection score  for $w$.
We then compute the precision, recall, $F1$, and accuracy scores, again using $\tau_p$ as above.

\section{Results and Discussion}

Table~\ref{tbl:oracle_evaluation} shows the localisation accuracies in the \textit{oracle} setting for both the PSC and GradCAM models when supervised with BoW labels and visual context. Table~\ref{tbl:actual_evaluation} gives the scores in the \textit{actual} setting, where detection is also taken into account, using a threshold of $\alpha = 0.4$.

\begin{table}[b]
	\centering
	\renewcommand{\arraystretch}{1.1}
	\begin{tabularx}{0.48\textwidth}{@{}lCC@{}} 
		\toprule
		& \multicolumn{2}{c}{Supervision method} \\
		\cmidrule(l){2-3}
		Mechanism & BoW & Visual \\
		\midrule
		\textbf{PSC} & $63.6$ & $19.1$\\
		\textbf{GradCAM} & $17.8$ & $16.0$\\
		\bottomrule
	\end{tabularx}
	\vspace{5pt}
	\caption{Oracle localisation accuracy ($\%$) when assuming perfect detection.}
	\label{tbl:oracle_evaluation}		
	
\end{table}

\begin{table}[!b]
	\renewcommand{\arraystretch}{1.1}
	\centering
	\begin{tabularx}{0.95\textwidth}{@{}lCCCcCCCc@{}} %
		\toprule
		& \multicolumn{4}{c}{BoW} & \multicolumn{4}{c}{Visual}\\
		\cmidrule(lr){2-5}\cmidrule(l){6-9}
		Mechanism & $P$ & $R$ & $F1$ & Accuracy & $P$ & $R$ & $F1$ & Accuracy\\
		\midrule
		\textbf{PSC} & $75.2$ & $53.0$ & $62.2$ & $50.4$ & $28.6$ & $8.0$ & $12.5$ & $7.6$\\
		\textbf{GradCAM} & $17.7$ & $24.5$ & $20.5$ & $13.2$ & $5.0$ & $5.7$ & $5.3$ & $4.4$\\
		\bottomrule
	\end{tabularx}
	\vspace{5pt}
	\caption{Actual localisation precision, recall, $F1$ and accuracy ($\%$) when taking detection into account with a threshold of $\alpha = 0.4$.} 
	\label{tbl:actual_evaluation}
\end{table}

\begin{figure}[t]
	\centering
	\includegraphics[width=0.6\linewidth]{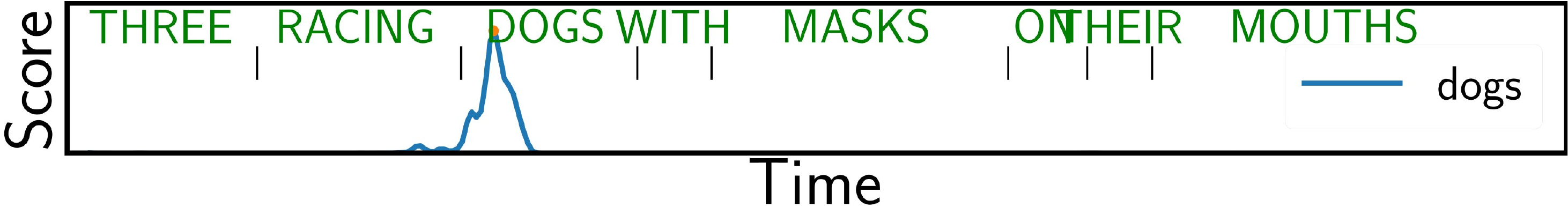} \\[-5pt]
	{\small (a)} \\[5pt]
	
	\includegraphics[width=0.6\linewidth]{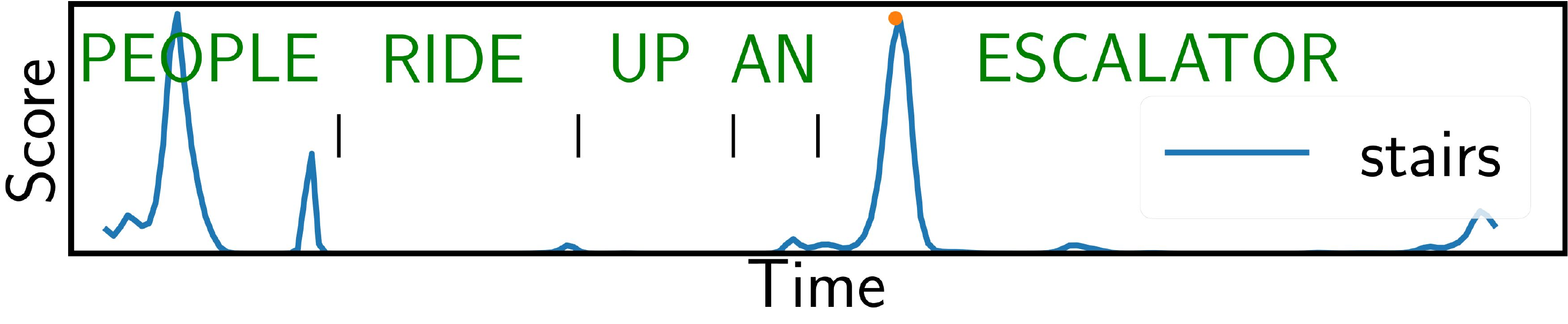} \\[-5pt]
	{\small (b)} \\[5pt]
	
	\includegraphics[width=0.6\linewidth]{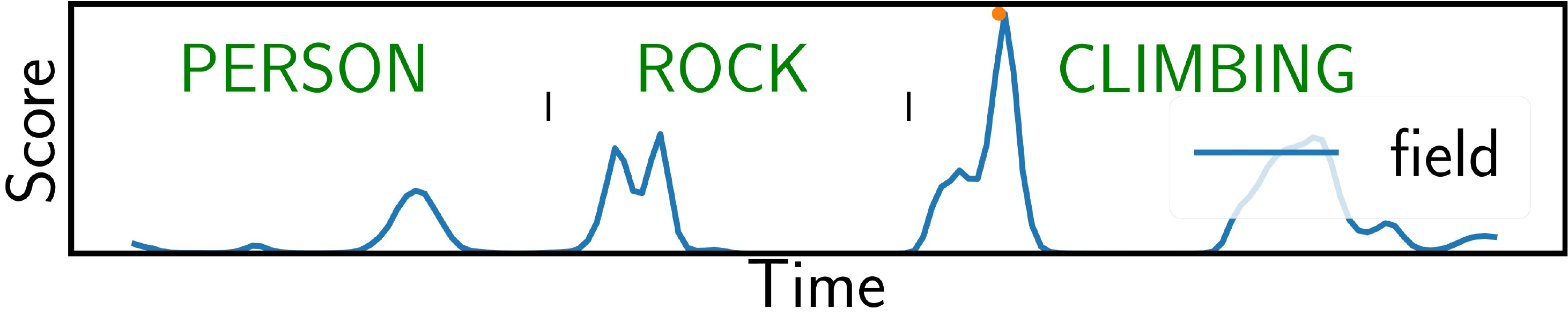} \\[-5pt]
	{\small (c)}
	\vspace{5pt}
	\caption{Examples of localisation with the visually supervised PSC model. The keyword being localised is shown on the right of each plot.}
	\label{fig:metric}
\end{figure}

Of the two forms of supervision, BoW labels leads to consistently better localisation 
than visual supervision. This is not surprising since 
different speakers could describe the same image in many different ways. Moreover, the visual tagger (which provides the training signal here) can assign high probabilities to concepts that no speaker would refer to (but which is nevertheless present in an image) or could tag semantically related words. As qualitative evidence of the last-mentioned, Figure \ref{fig:metric}(b) shows that ``escalator'' is localised when prompted with 
``stairs''. Despite many incorrect predictions when using visual supervision, e.g.\ Figure~\ref{fig:metric}(c), visual supervision does provide some signal for localisation, with the PSC model achieving a precision of almost 30\% in Table~\ref{tbl:actual_evaluation}. Figure~\ref{fig:metric}(a) shows a correct localisation. This is noteworthy since this model is trained without any textual labels and is still able to make text~predictions.

\begin{figure}[!t]
	\centering
	\includegraphics[width=0.8\columnwidth]{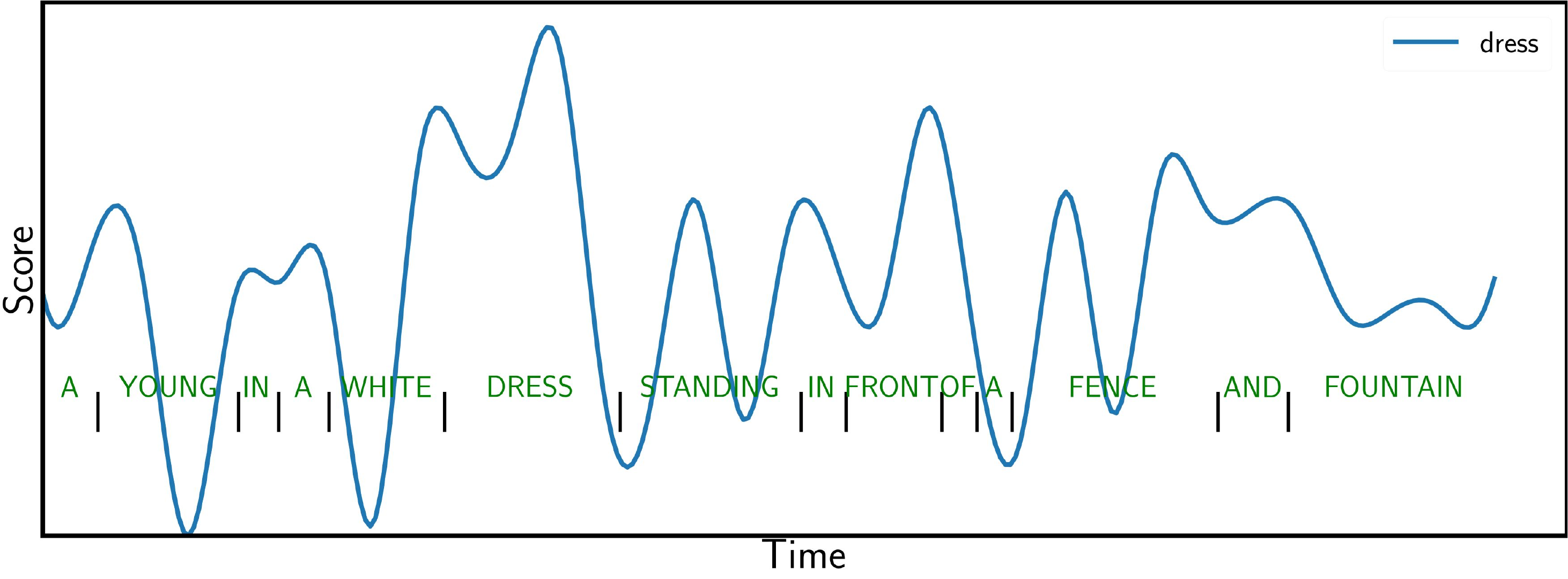}
	\vspace{5pt}
	\captionof{figure}{An example localisation with the GradCAM model for the keyword ``dress''.}
	\label{fig:gradcam_localisation}
\end{figure}

Of the two localisation mechanisms, PSC outperforms GradCAM on all metrics with both forms of supervision. Figure~\ref{fig:gradcam_localisation} gives an example of GradCAM localisation. We see that, in contrast to PSC (Figure~\ref{fig:metric}), peaks are produced on many words.We speculate that this is because GradCAM tries to identify the parts of the input that will cause a large change in a particular output unit.
In single-label multi-class classification, for which GradCAM was developed, a higher probability for a particular output implies lower probabilities for others. But this is not the case for multi-label classification, as used here, and the gradients of multiple words could therefore affect the output, and this would be captured in the gradients. Table~\ref{tbl:detection_evaluation} shows that when considering detection, i.e.\ only predicting whether a word is present in an utterance without evaluating localisation, the GradCAM model in fact outperforms the PSC model.

\begin{table*}[t]
	\centering
	\renewcommand{\arraystretch}{1.1}
   \begin{tabularx}{0.8\textwidth}{@{}lCCCCCC@{}} 
    \toprule
      & \multicolumn{3}{c}{$\alpha = 0.4$} & \multicolumn{3}{c}{$\alpha = 0.6$}\\
       \cmidrule(lr){2-4}\cmidrule(l){5-7}
      Model & $P$ & $R$ & $F1$ & $P$ & $R$ & $F1$\\
      \midrule
      \multicolumn{3}{@{}l}{\textit{Visual supervision:}}\\
      \textbf{PSC} & $44.5$ & $9.8$ & $16.1$ & $74.7$ & $4.3$ & $8.1$\\
      \textbf{GradCAM} & $29.3$ & $22.0$ & $25.1$ & $42.7$ & $12.7$ & $19.6$\\[2pt]
      \multicolumn{3}{@{}l}{\textit{BoW supervision:}}\\
      \textbf{PSC} & $82.2$ & $49.0$ & $61.4$ & $87.8$ & $46.1$ & $60.4$\\
      \textbf{GradCAM} & $79.3$ & $52.6$ & $63.2$ & $82.5$ & $50.9$ & $63.0$\\
      \bottomrule
    \end{tabularx}
	\vspace{5pt}
  \captionof{table}{Keyword detection scores (without considering localisation) with threshold $\alpha$.} 
\label{tbl:detection_evaluation}

\end{table*}

\section{Conclusion}

We asked whether keyword localisation in speech is possible with two forms of weak supervision when location information is not provided. We compared two localisation methods with two forms of supervision: bag-of-word (BoW) labels and visual context. While a saliency-based method performed poorly, a method where localisation is performed as part of the network performed well with BoW supervision and showed that visual supervision does provide signal for higher precision localisation. As far as we know, this is the first work to use these localisation mechanisms with visual supervision. \citet{harwath2018} did consider localisation of a word given a corresponding image, but this is different from our study where we locate a word in speech based on a given text label.

Our results suggests a mismatch between saliency-based localisation and the multi-label model used here, with a superior detection model 
performing poorly in localisation. This suggest that better localisation should be possible given a mechanism better aligned to the model and multi-label classification loss.

\section*{Acknowledgements}

This work is supported in part by the National Research Foundation of South Africa (grant number: 120409), a Google Faculty Award for the last author, and Google Africa PhD scholarships for the first two authors.

\bibliographystyle{abbrvnat}
\bibliography{refs}

\end{document}